\documentclass[runningheads]{llncs}
\usepackage{graphicx}
\usepackage{url}
\def\Fig#1{{Fig.\ \ref{fig:#1}}}
\def\Eq#1{{Eq.\ \ref{eq:#1}}}

\begin{document}
\title{MULAN: Multitask Universal Lesion Analysis Network for Joint Lesion Detection, Tagging, and Segmentation}
\titlerunning{MULAN: Multitask Universal Lesion Analysis Network}
%
\author{Ke Yan$ ^1 $ \and Youbao Tang$ ^1 $ \and Yifan Peng$ ^2 $ \and Veit Sandfort$^1$ \and \\ Mohammadhadi Bagheri$ ^1 $ \and Zhiyong Lu$ ^2 $ \and Ronald M. Summers$ ^1 $}

\authorrunning{K. Yan et al.}

\institute{Imaging Biomarkers and Computer-Aided Diagnosis Laboratory, Clinical Center \and
	National Center for Biotechnology Information, National Library of Medicine\\
	$ ^{1,2} $ National Institutes of Health, Bethesda, MD 20892\\
\email{yankethu@gmail.com, \{youbao.tang, yifan.peng, mohammad.bagheri, zhiyong.lu, rms\}@nih.gov, veit.sandfort@googlemail.com}}
\maketitle              
\begin{abstract}

When reading medical images such as a computed tomography (CT) scan, radiologists generally search across the image to find lesions, characterize and measure them, and then describe them in the radiological report. To automate this process, we propose a multitask universal lesion analysis network (MULAN) for joint detection, tagging, and segmentation of lesions in a variety of body parts,
which greatly extends existing work of single-task lesion analysis on specific body parts. MULAN is based on an improved Mask R-CNN framework with three head branches and a 3D feature fusion strategy. It achieves the state-of-the-art accuracy in the detection and tagging tasks on the DeepLesion dataset, which contains 32K lesions in the whole body. We also analyze the relationship between the three tasks and show that tag predictions can improve detection accuracy via a score refinement layer.

\end{abstract}

\section{Introduction}

Detection, classification, and measurement of clinically important findings (lesions) in medical images are primary tasks for radiologists \cite{Sahiner2018survey}. Generally, they search across the image to find lesions, and then characterize their locations, types, and related attributes to describe them in radiological reports. They may also need to measure the lesions, e.g., according to the RECIST guideline \cite{Eisenhauer2009RECIST}, for quantitative assessment and tracking. To reduce radiologists' burden and improve accuracy, there have been many efforts in the computer-aided diagnosis area to automate this process. For example, detection, attribute estimation, and malignancy prediction of lung nodules have been extensively studied \cite{Liao2019leaky,Wu2018joint}. Other works include detection and malignancy prediction of breast lesions \cite{Ribli2018breast}, classification of three types of liver lesions \cite{Diamant2016liver}, and segmentation of lymph nodes \cite{Tang2019LNGAN}. Variants of Faster R-CNN \cite{Ribli2018breast,Liao2019leaky} have been used for detection, whereas patch-based dictionaries \cite{Diamant2016liver} or networks \cite{Liao2019leaky,Wu2018joint} have been studied for classification and segmentation.

Most existing work on lesion analysis focused on certain body parts (lung, liver, etc.). In practice, a radiologist often needs to analyze various lesions in multiple organs. Our goal is to build such a universal lesion analysis algorithm to mimic radiologists, which to the best of our knowledge is the first work on this problem. To this end, we attempt to integrate the three tasks in one framework. Compared to solving each task separately, the joint framework will be not only more efficient to use, but also more accurate, since different tasks may be correlated and help each other \cite{Tang2019Uldor,Wu2018joint}.

We present the multitask universal lesion analysis network (MULAN) which can detect lesions in CT images, predict multiple tags for each lesion, and segment it as well. This end-to-end framework is based on an improved Mask R-CNN \cite{He2017MaskRCNN} with three branches: detection, tagging, and segmentation. The tagging (multilabel classification) branch learns from tags mined from radiological reports. We extracted 185 fine-grained and comprehensive tags describing the body part, type, and attributes of the lesions. The relation between the three tasks is analyzed by experiments in this paper. Intuitively, lesion detection can benefit from tagging, because the probability of a region being a lesion is associated with its attribute tags. We propose a score refinement layer in MULAN to explicitly fuse the detection and tagging results and improve the accuracy of both. A 3D feature fusion strategy is developed to leverage the 3D context information to improve detection accuracy. 

MULAN is evaluated on the DeepLesion \cite{Yan2018DeepLesion} dataset, a large-scale and diverse dataset containing measurements and 2D bounding-boxes of over 32K lesions from a variety of body parts on computed tomography (CT) images. It has been adopted to learn models for universal lesion detection \cite{Yan20183DCE,Tang2019Uldor}, measurement \cite{Tang2018RECIST}, and classification \cite{Yan2019Lesa}. On DeepLesion, MULAN achieves the state-of-the-art accuracy in detection and tagging and performs comparable in segmentation. It outperforms the previous best detection result by 10\%. We released the code of MULAN in \footnote[1]{\small \url{https://github.com/rsummers11/CADLab/tree/master/MULAN_universal_lesion_analysis}}.

\section{Method}
\begin{figure}[]
	\begin{center}
		\includegraphics[width=\linewidth,trim=60 160 70 155, clip]{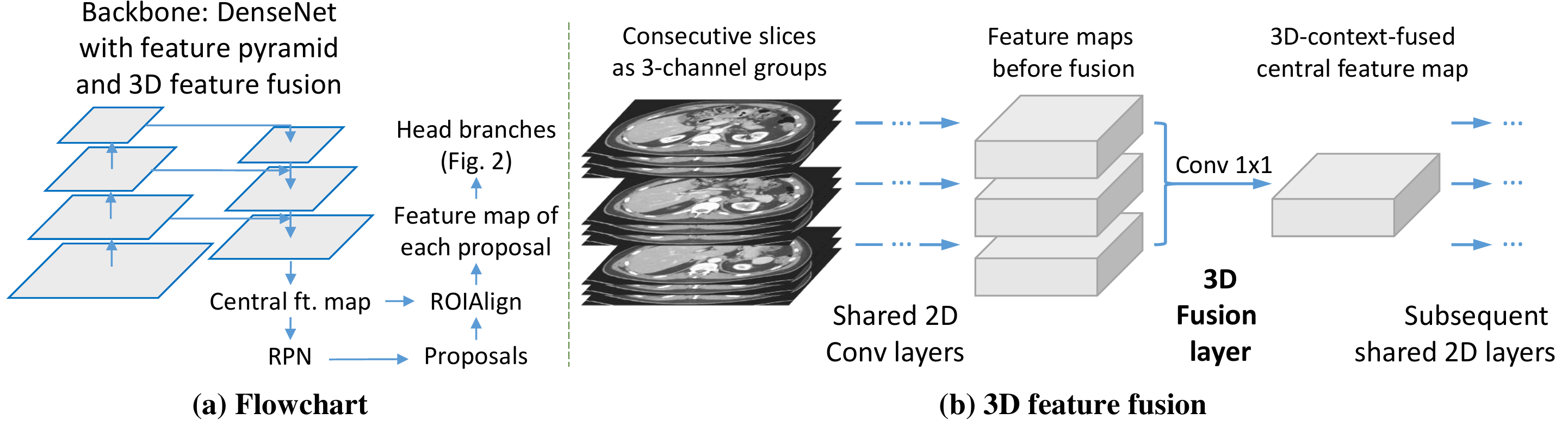} 
	\end{center}
	\caption{Flowchart of MULAN and the 3D feature fusion strategy.}
	\label{fig:net_backbone}
\end{figure}
The flowchart of the multitask universal lesion analysis network (MULAN) is displayed in \Fig{net_backbone} (a). Similar to Mask R-CNN \cite{He2017MaskRCNN}, MULAN has a backbone network to extract a feature map from the input image, which is then used in the region proposal network (RPN) to predict lesion proposals. Then, an ROIAlign layer \cite{He2017MaskRCNN} crops a small feature map for each proposal, which is used by three head branches to predict the lesion score, tags, and mask of the proposal.

\subsection{Backbone with 3D Feature Fusion}

A good backbone network is able to encode useful information of the input image into the feature map. In this study, we adopt the DenseNet-121 \cite{Huang2017DenseNet} in the backbone with the last dense block and transition layer removed, as we found removing them slightly improved accuracy and speed. Next, we employ the feature pyramid strategy \cite{Lin2016Pyramid} to add fine-level details into the feature map. 
This strategy also increases the size of the final feature map, which will benefit the detection and segmentation of small lesions. Different from the original feature pyramid network \cite{Lin2016Pyramid} which attaches head branches to each level of the pyramid, we attach the head branches only to the finest level \cite{Liao2019leaky,Wu2018joint}.

3D context information is very important when differentiating lesions from non-lesions \cite{Yan20183DCE}. 3D CNNs have been used for lung nodule detection \cite{Liao2019leaky}. However, they are memory-consuming, thus smaller networks need to be used. Universal lesion detection is much more difficult than lung nodule detection, so networks with more channels and layers are potentially desirable. Yan et al. \cite{Yan20183DCE} proposed 3D context enhanced region-based CNN (3DCE) and achieved better detection accuracy than a 3D CNN in the DeepLesion dataset. They first group consecutive axial slices in a CT volume into 3-channel images. The upper and lower images provide 3D context for the central image. A feature map is then extracted for each image with a shared 2D CNN. Lastly, they fuse the feature maps of all images with a convolutional (Conv) layer to produce the 3D-context-enhanced feature map for the central image and predict 2D boxes for the lesions on it.

The drawback of 3DCE is that the 3D context information is fused only in the last Conv layer, which limits the network's ability to learn more complex 3D features. As shown in \Fig{net_backbone} (b), we improve 3DCE to relieve this issue. The basic idea is to fuse features of multiple slices in earlier Conv layers. Similar to 3DCE, feature maps (FMs) are fused with a Conv layer (i.e., the 3D fusion layer). Then, the fused central FM is used to replace the original central FM, while the upper and lower FMs are kept unchanged. All FMs are then fed to subsequent Conv layers. Because the new central FM contains 3D context information, sophisticated 3D features can be learned in subsequent layers with nonlinearity. This 3D fusion layer can be inserted between any two layers of the original 2D CNN.  In MULAN, one 3D fusion layer is inserted after dense block 2 and another one after the last layer of the feature pyramid. We found fusing 3D context in the beginning of the CNN (before dense block 2) is not good possibly because the CNN has not yet learned good semantic 2D features by then. At the end of the network, only the central feature map is used as the FM of the central image.

\subsection{Head Branches and Score Refinement Layer}

\begin{figure}[]
	\begin{center}
		\includegraphics[width=\linewidth,trim=170 160 170 185, clip]{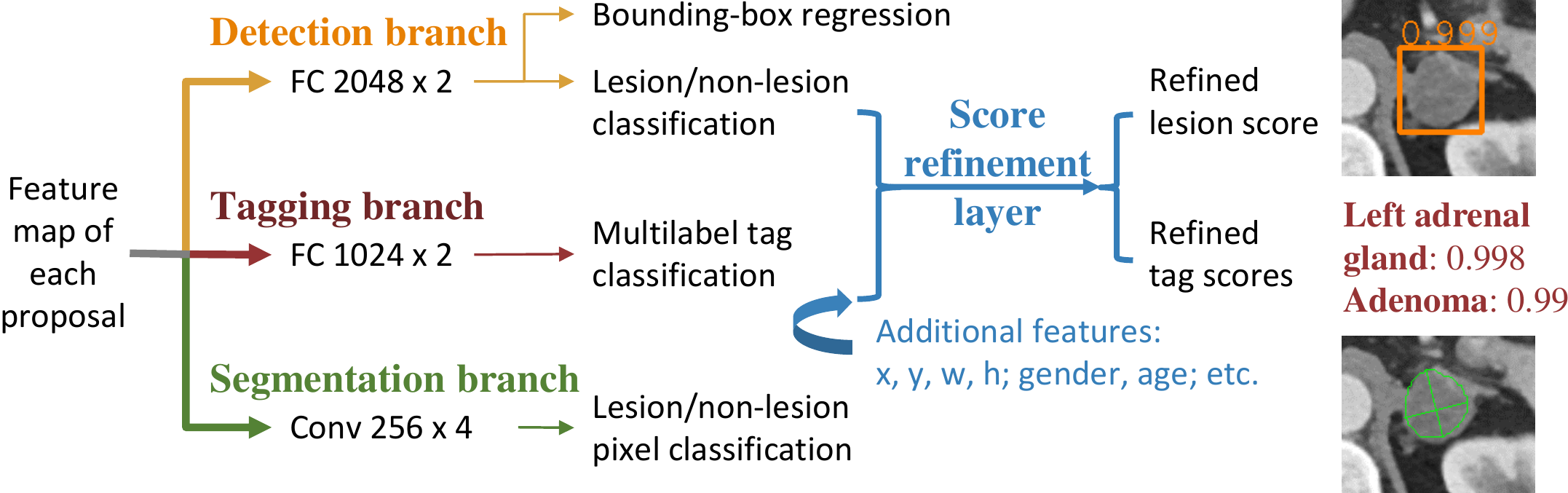} 
	\end{center}
	\caption{Illustration of the head branches and the score refinement layer of MULAN.}
	\label{fig:net_heads}
\end{figure}

The structure and function of the three head branches are shown in \Fig{net_heads}. The \textbf{detection branch} consists of two 2048D fully connected layers (FC) and predicts the lesion score of each proposal, i.e., the probability of the proposal being a lesion. It also conducts bounding-box regression to refine the box \cite{He2017MaskRCNN}.

The \textbf{tagging branch} predicts the body part, type, and attributes (intensity, shape, etc.) of the lesion proposal. It applies the same label mining strategy as that in LesaNet \cite{Yan2019Lesa}. We first construct the lesion ontology based on the RadLex lexicon. To mine training labels, we tokenize the sentences in the radiological reports of DeepLesion, and then match and filter the tags in the sentences using a text mining module. 185 tags with more than 30 occurrences in DeepLesion are kept. A weighted binary cross-entropy loss is applied on each tag. The hierarchical and mutually exclusive relations between the tags were leveraged in a label expansion strategy and a relational hard example mining loss to improve accuracy \cite{Yan2019Lesa}. The score propagation layer and the triplet loss in \cite{Yan2019Lesa} are not used. Due to space constraints, we refer readers to the supplementary material (sup.\ mat.) and \cite{Yan2019Lesa} for more implementation details in this branch.

For the \textbf{segmentation branch}, we follow the method in \cite{Tang2019Uldor} and generate pseudo-masks of lesions for training. The DeepLesion dataset does not contain lesions' ground-truth masks. Instead, each lesion has a RECIST measurement \cite{Eisenhauer2009RECIST}, namely a long axis and a short axis annotated by radiologists. They are utilized to generate four quadrants as the estimation of the real mask \cite{Tang2019Uldor}, since most lesions have ellipse-like shapes. We use the Dice loss \cite{Wu2018joint} as it works well in balancing foreground and background pixels. The predicted mask can be easily used to compute the contour and then the RECIST measurement of the lesion, see \Fig{net_heads} for an example.

Intuitively, detection (lesion/non-lesion classification) is closely related to tagging. One way to exploit their synergy is to combine them in one branch to make them share FC features. However, this strategy led to inferior accuracy for both tasks in our experiments probably because detecting a variety of lesions is a hard problem and requires rich features with high nonlinearity, thus a dedicated branch is necessary. In this study, we propose to combine them at the decision level. Specifically, for each lesion proposal, we join its lesion score from the detection branch and the 185 tag scores from the tagging branch as a feature vector, then predict the lesion and tag scores again using a \textbf{score refinement layer} (SRL). Tag predictions can thus support detection explicitly. We also add new features as the input of the layer including the statistics of the proposal ($ x,y, $ width, height), the patient's gender, and age. Other relevant features such as medical history and lab results may also be considered. In MULAN, SRL is a simple FC layer as we found more nonlinearity did not improve results possibly due to overfitting. The losses for detection and tagging after this layer are the same as those in the respective branches.

More implementation details of MULAN are depicted in the sup.\ mat.

\section{Experiments and Discussion}

\textbf{Implementation:} MULAN was implemented in PyTorch based on the maskrcnn-benchmark\footnote[2]{\url{https://github.com/facebookresearch/maskrcnn-benchmark}} project. The DenseNet backbone was initialized with an ImageNet pretrained model. The score refinement layer was initialized with an identity matrix so that the scores before and after it were the same when training started. Other layers were randomly initialized. Each mini-batch had 8 samples, where each sample consisted of three 3-channel images for 3D fusion (\Fig{net_backbone}). We used SGD to train MULAN for 8 epochs and set the base learning rate to 0.004, then reduced it by a factor of 10 after the 4th and 6th epochs. It takes MULAN 30ms to predict a sample during inference on a Tesla V100 GPU.

\textbf{Data:} The DeepLesion dataset \cite{Yan2018DeepLesion} contains 32,735 lesions and was divided into training (70\%), validation (15\%), and test (15\%) sets at the patient level. When training, we did data augmentation for each image in three ways: random resizing with a ratio of 0.8$ \sim $1.2; random translation of -8$ \sim $8 pixels in $ x $ and $ y $ axes; and 3D augmentation. A lesion in DeepLesion was annotated in one axial slice, but the actual lesion also exists in approximately the same position in several neighboring slices depending on its diameter and the slice interval. Therefore, we can do 3D augmentation by randomly shifting the slice index within half of the lesion's short diameter. Each of these three augmentation methods improved detection accuracy by 0.2$ \sim $0.4\%. Some examples of DeepLesion are presented in Section 1 of the sup.\ mat.

\textbf{Metrics:} For detection, we compute the sensitivities at 0.5, 1, 2, and 4 false positives (FPs) per image \cite{Yan20183DCE} and average them, which is similar to the evaluation metric of the LUNA dataset \cite{Liao2019leaky}. For tagging, we use the 500 manually tagged lesions in \cite{Yan2019Lesa} for evaluation. The area under the ROC curve (AUC) and F1 score are computed for each tag and then averaged. Since there are no ground-truth (GT) masks in DeepLesion except for RECIST measurements \cite{Tang2018RECIST}, we use the average distance from the endpoints of the GT measurement to the predicted contour as a surrogate criterion (see sup.\ mat.\ Section 2). The second criterion is the average error of length of the estimated RECIST diameters, which are very useful values for radiologists and clinicians \cite{Eisenhauer2009RECIST}.

\begin{figure}[]
	\begin{center}
		\includegraphics[width=.8\linewidth,trim=95 320 98 80, clip]{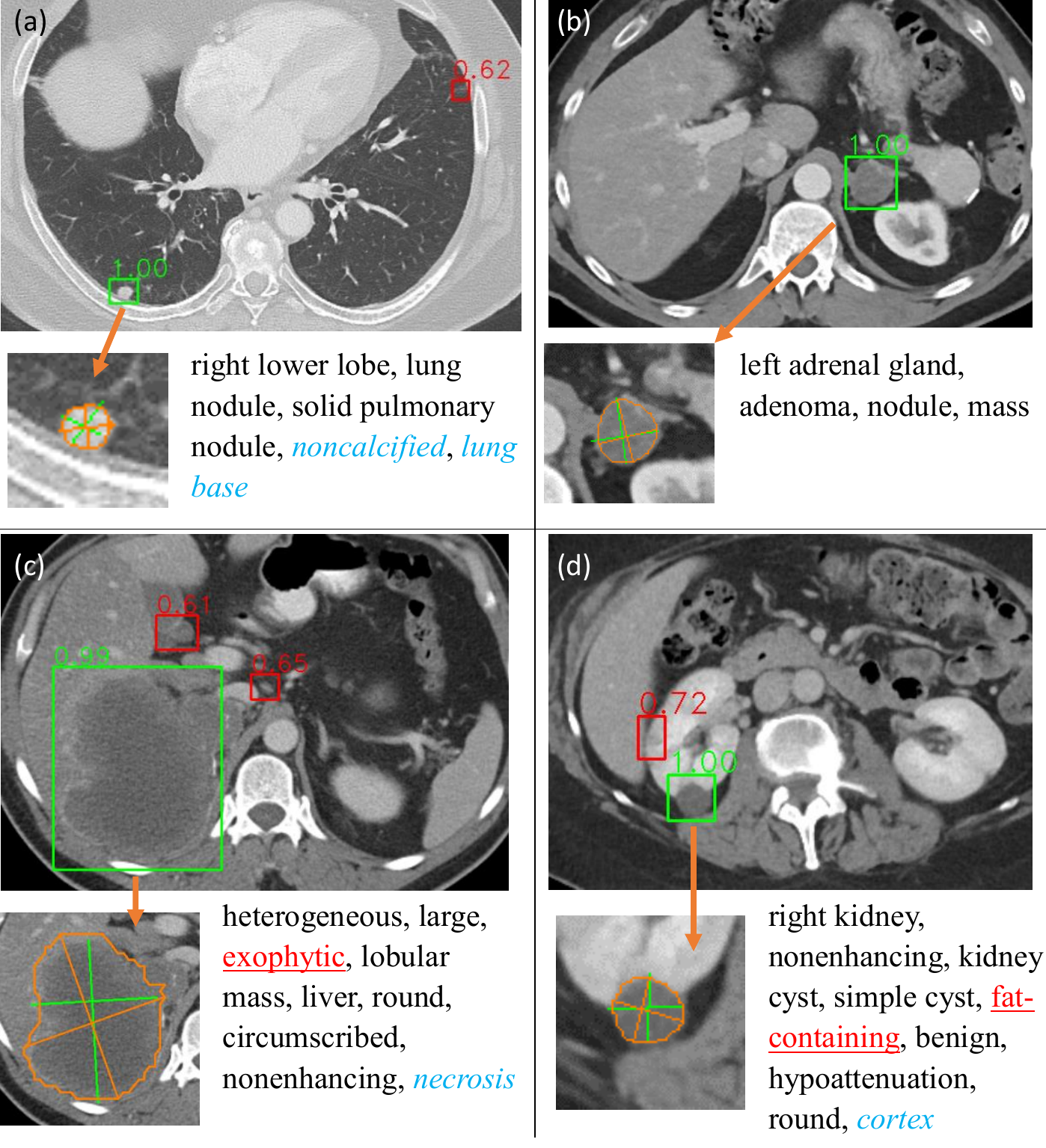} 
	\end{center}
	\caption{Examples of MULAN's lesion detection, tagging, and segmentation results on the test set of DeepLesion. For detection, boxes in green and red are predicted TPs and FPs, respectively. The number above each box is the lesion score (confidence). For tagging, tags in black, red (underlined), and blue (italic) are predicted TPs, FPs, and FNs, respectively. They are ranked by their scores. For segmentation, the green lines are ground-truth RECIST measurements; the orange contours and lines show predicted masks and RECIST measurements, respectively. More visual examples are provided in sup. mat. Section 3. (TP: true positive; FP: false positive; FN: false negative)}
	\label{fig:result_example}
\end{figure}

Qualitative and quantitative results are presented in \Fig{result_example} and Table \ref{tbl:res}, respectively. Note that in Table \ref{tbl:res}, tagging and segmentation accuracy were calculated by predicting tags and masks based on GT bounding-boxes, so that they were under the same setting as previous studies \cite{Tang2018RECIST,Yan2019Lesa} and independent of the detection accuracy. We will discuss the results of each task below.

\begin{table}[]
	\centering
	\setlength{\tabcolsep}{2.6pt}
	\renewcommand{\arraystretch}{1.2}
	\caption{Accuracy comparison and ablation studies on the test set of DeepLesion. 
	Bold results are the best ones. Underlined results in the ablation studies are the worst ones, indicating the ablated strategy is the most important for the criterion.}
	\begin{tabular}{lccccccc} 
		
		\hline 
		&	\multicolumn{1}{c}{Detection (\%)} && \multicolumn{2}{c}{Tagging (\%)} && \multicolumn{2}{c}{Segmentation (mm)} \\
		\cline{2-8}
			&	Avg. sensitivity	&&	AUC	&	F1	&& Distance	&	Diam. err.	\\
		\hline
		ULDor \cite{Tang2019Uldor}	& 69.22	&& --	& --	&& --	& -- \\
		3DCE \cite{Yan20183DCE}	& 75.55	&& --	& --	&& --	& -- \\
		LesaNet \cite{Yan2019Lesa} (rerun)	& --	&& 95.12	& 43.17	&& --	& -- \\
		Auto RECIST \cite{Tang2018RECIST} & --	&& --	& --	&& --	& \bf 1.7088 \\
		MULAN	& \bf 86.12	&& \bf 96.01  & \bf 45.53	&& 1.4138 & 1.9660 \\
		\hline
		(a) w/o feature pyramid	& 79.73	&& 95.51  & 43.44	&& \underline{1.6634}	& \underline{2.3780} \\
		(b) w/o 3D fusion	& \underline{79.57}	&& 95.88	& 44.28	&& 1.4120	& 1.9756 \\
		(c) w/o detection branch	& --	&& \underline{95.16}  & \underline{40.03}	&& \bf 1.2445	& 1.7837 \\
		(d) w/o tagging branch	& 84.79	&& --	& --	&& 1.4230	& 1.9589 \\
		(e) w/o mask branch	& 85.21	&& 95.87  & 43.76 && --	& -- \\
		(f) w/o score refine. layer	& 84.24	&& 95.65	& 44.59	&& 1.4260	& 1.9687 \\
		
		\hline
	\end{tabular}
	\label{tbl:res} 
\end{table}


\textbf{Detection:} Table \ref{tbl:res} shows that MULAN significantly surpasses existing work on universal lesion detection by over 10\% in average sensitivity. According to the ablation study, 3D fusion and feature pyramid improve detection accuracy the most. If the tagging branch is not added (ablation study (d)), the detection accuracy is 84.79\%; When it is added, the accuracy slightly drops to 84.24\% (ablation study (f)). However, when the score refinement layer (SRL) is added, we achieve the best detection accuracy of 86.12\%. We hypothesize that SRL effectively exploits the correlation between the two tasks and uses the tag predictions to refine the lesion detection score. To verify the impact of SRL, we randomly re-split the training and validation set of DeepLesion five times and found MULAN with SRL always outperformed it without SRL by $ 0.7\sim1.1\% $.

Examples in \Fig{result_example} show that MULAN is able to detect true lesions with high confidence score, although there are still FPs when normal tissues have a similar appearance with lesions. We analyzed the detection accuracy by tags and found lung masses/nodules, mediastinal and pelvic lymph nodes, adrenal and liver lesions are among the lesions with the highest sensitivity, while lesions in pancreas, bone, thyroid, and extremity are relatively hard to detect. These conclusions can guide us to collect more training samples with the difficult tags in the future.

\textbf{Tagging:} MULAN outperforms LesaNet \cite{Yan2019Lesa}, a multilabel CNN designed for universal lesion tagging. According to ablation study (c), adding the detection branch improves tagging accuracy. This is probably because detection is hard and requires comprehensive features to be learned in the backbone of MULAN, which are also useful for tagging. \Fig{result_example} shows that MULAN is able to predict the body part, type, and attributes of lesions with high accuracy.

\textbf{Segmentation:} Our predicted RECIST diameters have an average error of 1.97mm compared with the GT diameters. From \Fig{result_example}, we can find that MULAN performs relatively well on lesions with clear borders, but struggles on those with indistinct or irregular borders, e.g., the liver mass in \Fig{result_example} (c). Ablation studies show that feature pyramid is the most crucial strategy. Another interesting finding is that removing the detection branch (ablation study (c)) markedly improves segmentation accuracy. The detection task impairs segmentation, which could be a major reason why the multitask MULAN cannot beat Auto RECIST \cite{Tang2018RECIST}, a framework dedicated to lesion measurement. It implies that better segmentation results may be achieved using a single-task CNN.

More detailed results are shown in supplementary material.

\section{Conclusion and Future Work}

In this paper, we proposed MULAN, the first multitask universal lesion analysis network which can simultaneously detect, tag, and segment lesions in a variety of body parts. The training data of MULAN can be mined from radiologists' routine annotations and reports with minimum manual effort \cite{Yan2018DeepLesion}. An effective 3D feature fusion strategy was developed. We also analyzed the interaction between the three tasks and discovered that: 1) Tag predictions could improve detection accuracy via a score refinement layer; 2) The detection task improved tagging accuracy but impaired segmentation performance.

Universal lesion analysis is a challenging task partially because of the large variance of appearances of the normal and abnormal tissues. Therefore, the 22K training lesions in DeepLesion are still not sufficient for MULAN to learn, which is a main reason for its FPs and FNs. In the future, more training data need to be mined. We also plan to apply or finetune MULAN on other applications of specific lesions. We hope MULAN can be a useful tool for researchers focusing on different types of lesions.

{\small \textbf{Acknowledgments:} This research was supported by the Intramural Research Programs of the National Institutes of Health (NIH) Clinical Center and National Library of Medicine (NLM). It was also supported by NLM of NIH under award number K99LM013001. We thank NVIDIA for GPU card donations.}

\section{Appendix}

\subsection{Introduction to the DeepLesion Dataset}
\begin{figure}[]
	\begin{center}
		\includegraphics[width=\linewidth,trim=90 260 70 80, clip]{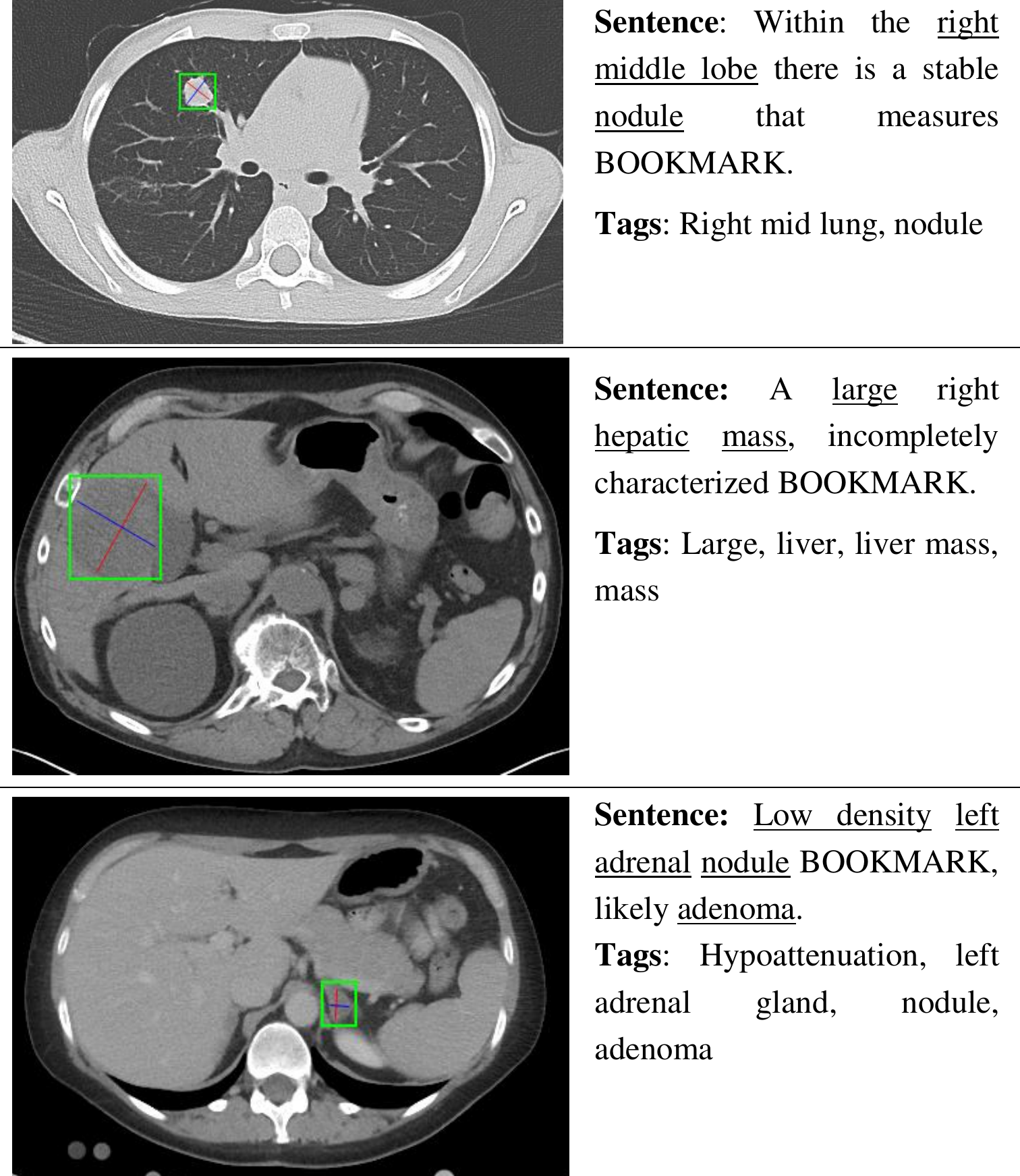} 
	\end{center}
	\caption{Examples of the CT images, annotations, and reports in DeepLesion \cite{Yan2018DeepLesion}. The red and blue lines in the images are the RECIST measurements. The green boxes are the bounding-boxes. The sentences are extracted from radiological reports according to the bookmarks \cite{Yan2019Lesa}. The tags are mined from the sentences and normalized \cite{Yan2019Lesa}.}
	\label{fig:dataset_example}
\end{figure}

The DeepLesion dataset \cite{Yan2018DeepLesion} was mined from a hospital's picture archiving and communication system (PACS) based on bookmarks, which are markers annotated by radiologists during their routine work to measure significant image findings. It is a large-scale dataset with 32,735 lesions on 32,120 axial slices from 10,594 CT studies of 4,427 unique patients. There are 1 -- 3 lesions in each axial slice. Different from existing datasets that typically focus on one type of lesion, DeepLesion contains a variety of lesions including those in lungs, livers, kidneys, etc., and enlarged lymph nodes.

Each lesion in DeepLesion has a RECIST measurement \cite{Eisenhauer2009RECIST}, which consists of two lines: one measuring the longest diameter of the lesion and the second measuring its longest perpendicular diameter in the axial plane, see \Fig{dataset_example}. From these two diameters, we can compute a 2D bounding box to train a lesion \textbf{detection} algorithm \cite{Yan2018DeepLesion}, as well as generate a psuedo-mask to train a lesion \textbf{segmentation} algorithm \cite{Tang2019Uldor}.

Besides measuring the lesions, radiologists often describe them in radiological reports and use a hyperlink (shown as ``BOOKMARK'' in \Fig{dataset_example}) to link the measurement with the sentence. We can extract tags that describe the lesion in the sentence to train a lesion \textbf{tagging} algorithm \cite{Yan2019Lesa}. The predicted tags can provide comprehensive and fine-grained semantic information for the user to understand the lesion.

\subsection{Additional Details in Methods}

\subsubsection{Backbone}

The backbone structure of MULAN is a truncated DenseNet-121 \cite{Huang2017DenseNet} (87 Conv layers after truncation) with feature pyramid \cite{Lin2016Pyramid} and 3D feature fusion. The finest level of the feature pyramid corresponds to dense block 1 and has stride 4 \cite{Lin2016Pyramid}. The channel number after feature pyramid is 512.

\subsubsection{Detection and Segmentation}

The structures of the region proposal network (RPN), detection branch, and mask branch are similar to those in Mask R-CNN \cite{He2017MaskRCNN}. Five anchor scales (16, 24, 32, 48, 96) and three anchor ratios (1:2, 1:1, 2:1) are used in RPN. The loss function for detection and segmentation is
\begin{equation}\label{eq:loss_det_seg}
L_{\rm det,seg} = L_{\rm RPN,cls} + L_{\rm {RPN,box}} + L_{\rm {det,cls}} + 10L_{\rm {det,box}} + L_{\rm {seg,dice}},
\end{equation}
where $ L_{\rm RPN,cls} $ and $ L_{\rm {RPN,box}} $ are the classification (lesion vs.\ non-lesion) and bounding-box regression \cite{He2017MaskRCNN} losses of RPN; $ L_{\rm {det,cls}} $ and $ L_{\rm {det,box}} $ are those in the detection branch; $ L_{\rm {seg,dice}} $ is the Dice loss \cite{Milletari2016Vnet} in the segmentation branch.

\subsubsection{Tagging}

The label mining strategy and the loss function of the tagging branch are similar to \cite{Yan2019Lesa}, except that the score propagation layer and the triplet loss are not used. Based on the RadLex lexicon \cite{Langlotz2006RadLex}, we run whole-word matching in the sentences to extract the lesion tags and combine all synonyms. Some tags in the sentence are not related to the lesion in the image, so we use a text-mining module \cite{Yan2019Lesa} to filter the irrelevant tags. The final 185 tags can be categorized into three classes \cite{Yan2019Lesa}: {\bf 1.\ Body parts}, which include coarse-level body parts (e.g., chest, abdomen), organs (lung, lymph node), fine-grained organ parts (right lower lobe, pretracheal lymph node), and other body regions (porta hepatis, paraspinal); {\bf 2.\ Types}, which include general terms (nodule, mass) and more specific ones (adenoma, liver mass); and {\bf 3.\ Attributes}, which describe the intensity, shape, size, etc., of the lesions (hypoattenuation, spiculated, large).

The tagging branch predicts a score $ s_{i,c} $ for each tag $ c $ of each proposal $ i $. Because positive labels are sparse for most tags, we adopt a \textbf{weighted cross-entropy} (WCE) loss \cite{Yan2019Lesa} for each tag as in \Eq{loss_wce}, where $ B $ is the number of true lesions in a minibatch, $ C $ is the number of tags; $ \sigma_{i,c} = {\rm sigmoid}(s_{i,c})$; $ y_{i,c} \in\{0,1\}$ is the ground-truth of lesion $ i $ having tag $ c $. The loss weights are $ \beta_c^{\rm p}=|P_c+N_c|/|2P_c|, \beta_c^{\rm n}=|P_c+N_c|/|2N_c| $, $ P_c,N_c $ are the number of positive and negative labels of tag $ c $ in the training set of DeepLesion, respectively. Similar to the segmentation branch, the tagging branch only considers proposals corresponding to true lesions in the loss function, since we do not know the ground-truth tags of non-lesions, although non-lesions can also have body parts and attributes.
\begin{equation}\label{eq:loss_wce}
L_{{\rm tag,WCE}} = \sum_{i=1}^B\sum_{c=1}^C \left(\beta_c^{\rm p} y_{i,c} 
\log\sigma_{i,c} + \beta_c^{\rm n}(1-y_{i,c}) \log(1-\sigma_{i,c}) \right).
\end{equation}

There are hierarchical and mutually exclusive relations between the tags. For example, lung is the parent of left lung (if a lesion is in the left lung, it must be in the lung), while left lung and right lung are exclusive (they cannot both be true for one lesion). These relations can be leveraged to improve tagging accuracy. Tags extracted from reports are often not complete since radiologists typically do not write down all possible characteristics. If a tag is not mentioned in the report, it may still be true. To deal with this label noise problem, first, we use the \textbf{label expansion} strategy \cite{Yan2019Lesa} to infer the missing parent tags. If a child tag is mined from the report, all its parents will be set as true. Second, we use the \textbf{relational hard example mining} (RHEM) strategy \cite{Yan2019Lesa} to suppress reliable negative tags. If a tag is true, all its exclusive tags must be false, so we can define a new loss term $ L_{\rm tag,RHEM} $ to assign higher weights to these exclusive tags.

\subsubsection{Overall}

The overall loss function of MULAN is
\begin{equation}\label{eq:final_loss}
L = L_{{\rm det,seg}} + L_{{\rm tag,WCE}} + L_{{\rm tag,RHEM}} + L_{{\rm cls,SRL}} + L_{{\rm tag,WCE,SRL}},
\end{equation}
where $ L_{{\rm det,seg}} $ is defined in \Eq{loss_det_seg}, and $ L_{{\rm tag,WCE}} $ and $ L_{{\rm tag,RHEM}} $ are the losses of the tagging branch. $ L_{{\rm cls,SRL}} $ and $ L_{{\rm tag,WCE,SRL}} $ are the losses of the score refinement layer, which have the same forms as $ L_{\rm {det,cls}} $ in \Eq{loss_det_seg} and $ L_{{\rm tag,WCE}} $ in \Eq{loss_wce}, respectively.

\subsection{Additional Details in Experiments and Results}

\subsubsection{Image Preprocessing Method}

We rescaled the 12-bit CT intensity range to floating-point numbers in [0,255] using a single windowing (-1024--3071 HU) that covers the intensity ranges of the lung, soft tissue, and bone. Every image slice was resized so that each pixel corresponds to 0.8mm. The slice intervals of most CT scans in the dataset are either 1mm or 5mm. We interpolated in the $ z $-axis to make the intervals of all volumes 2mm. The black borders in images were clipped for computation efficiency. We used the official data split of DeepLesion. The input of our experiments are 9-slice sub-volumes in DeepLesion, including the key slice that contains the lesion, 4 slices superior to it, and 4 slices inferior \cite{Yan20183DCE}.

\subsubsection{Surrogate Evaluation Criteria for Lesion Segmentation}

\begin{figure}[]
	\begin{center}
		\includegraphics[trim=0 0 0 0, clip]{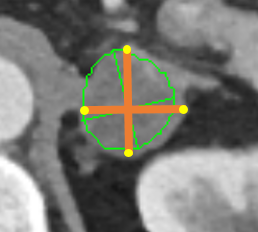} 
	\end{center}
	\caption{Illustration of the predicted mask (green contour), estimated RECIST measurement (green segments), and ground-truth RECIST measurement (orange segments with yellow endpoints) of a lesion.}
	\label{fig:seg_metric}
\end{figure}

There are no ground-truth (GT) masks in DeepLesion. Instead, each lesion has a GT RECIST measurement, so we use two surrogate metrics to evaluate segmentation results, see \Fig{seg_metric}. First, if the predicted mask is accurate, the endpoints of the GT RECIST measurement should be on its contour. Therefore, the average distance from the endpoint of the GT measurement to the contour of the predicted mask is a useful metric (the smaller the better). Second, if the predicted mask is accurate, the lengths of the estimated RECIST measurement (diameters) should be the same with the GT diameters. The average error of lengths is thus another useful metric (the smaller the better).

RECIST measurements \cite{Tang2018RECIST} can be easily estimated from the predicted mask. We first compute the contour of the predicted mask, then find two points on the contour with the largest distance to form the long axis. Next, we search on the contour to find the short axis that is perpendicular to the long axis and has the largest length.

\subsubsection{Additional Results}

\begin{figure}[]
	\begin{center}
		\includegraphics[width=\linewidth,trim=95 270 95 80, clip]{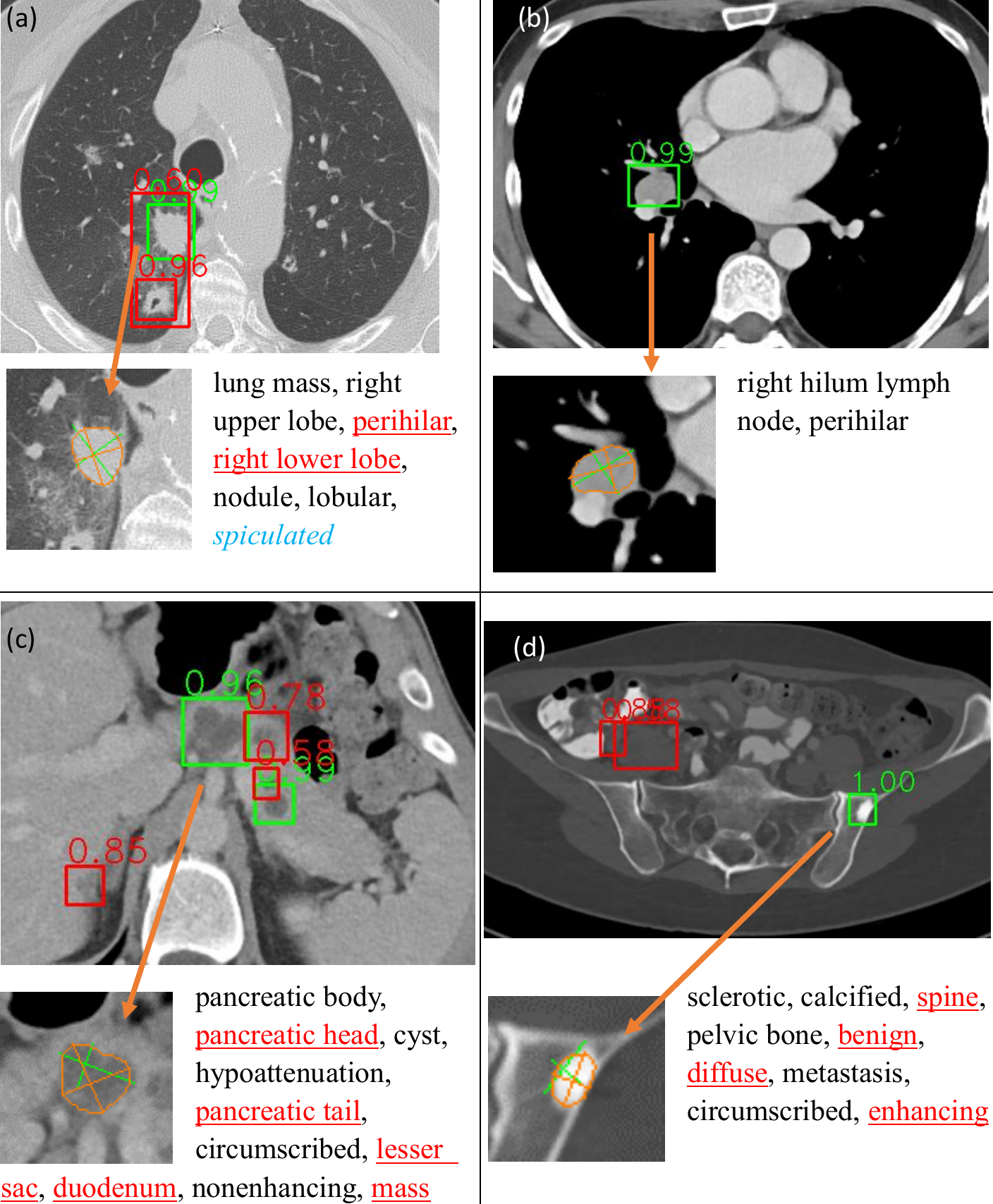} 
	\end{center}
	\caption{Examples of MULAN's lesion detection, tagging, and segmentation results on the test set of DeepLesion. For detection, boxes in green and red are predicted TPs and FPs, respectively. The number above each box is the lesion score (confidence). For tagging, tags in black, red (underlined), and blue (italic) are predicted TPs, FPs, and FNs, respectively. They are ranked according to their scores. For segmentation, the green lines are ground-truth RECIST measurements; the orange contours and lines show predicted masks and RECIST measurements, respectively. (TP: true positive; FP: false positive; FN: false negative)}
	\label{fig:sup_result_example}
\end{figure}

\begin{figure}[]
	\begin{center}
		\includegraphics[width=.7\linewidth,trim=0 0 0 0, clip]{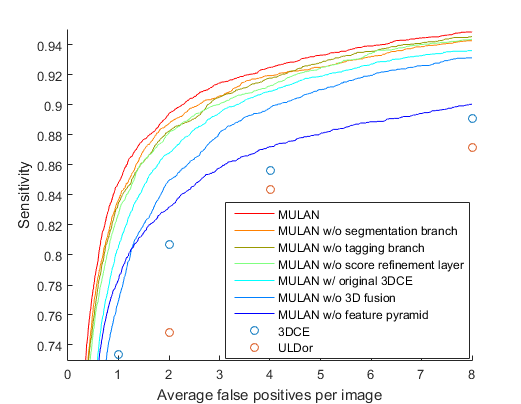} 
	\end{center}
	\caption{Free-response receiver operating characteristic (FROC) curve of various methods and variations of MULAN on the test set of DeepLesion.}
	\label{fig:FROC}
\end{figure}

We show the free-response receiver operating characteristic (FROC) curves in \Fig{FROC}. They correspond to the \textbf{detection} accuracies in Table 1 of the main paper. MULAN outperforms previous methods 3DCE and ULDor. In \Fig{sup_result_example}, the threshold for lesion scores is 0.5. Note that some FPs in the detection results are actually TPs, because there are missing lesion annotations in the test set of DeepLesion. Some examples include the smaller lung mass in \Fig{sup_result_example} (a) and the two smaller pancreatic lesions in \Fig{sup_result_example} (c).


To turn \textbf{tag} scores into decisions, we calibrated a threshold for each tag that yielded the best F1 on the validation set, and then applied it on the test set. In \Fig{sup_result_example}, MULAN is able to predict the body part, type, and attributes of lesions with high accuracy. Possible reasons of tagging errors include:
\begin{itemize}
	\item Some attributes with variable appearances and few training samples have some FPs, such as ``benign'' and ``diffuse'' in \Fig{sup_result_example} (d);
	\item Adjacent body parts may be confused by the model, such as ``right lower lobe'' in (a) and ``pancreatic head'',	``pancreatic tail'', ``lesser sac'', and ``duodenum'' in (c).
\end{itemize}


The threshold for \textbf{mask} prediction is 0.5. In \Fig{sup_result_example}, MULAN performs relatively well on lesions with clear borders ((a) and (b)), but struggles on those with indistinct borders ((c) and (d)). For the latter case, GT measurements may sometimes be inaccurate or not consistent (different radiologists have different opinions).

\subsection{Results using the Released Tags \cite{Yan2019Lesa} of DeepLesion}

In this paper, we used an NLP algorithm slightly different from \cite{Yan2019Lesa} to mine tags from reports. There are 185 tags in this paper and 171 in \cite{Yan2019Lesa}. Since the 171 tags of \cite{Yan2019Lesa} have been released in \footnote[3]{\url{https://github.com/rsummers11/CADLab/tree/master/LesaNet}}, we also retrained MULAN on these 171 tags so that the results can be compared with others' methods trained on the 171 tags. The results are shown in Table \ref{tbl:res_171tags} below. The definition of the metrics are the same with those in Table 1 of the main paper. The results are also similar with those in Table 1 of the main paper.

\begin{table}[]
	\centering
	\setlength{\tabcolsep}{2.6pt}
	\renewcommand{\arraystretch}{1.2}
	\caption{Accuracy comparison on the test set of DeepLesion using the 171 training tags of \cite{Yan2019Lesa}. Note that only LesaNet and MULAN used the tags.}
	\begin{tabular}{lccccccc}
		
		\hline 
		&	\multicolumn{1}{c}{Detection (\%)} && \multicolumn{2}{c}{Tagging (\%)} && \multicolumn{2}{c}{Segmentation (mm)} \\
		\cline{2-8}
		&	Avg. sensitivity	&&	AUC	&	F1	&& Distance	&	Diam. err.	\\
		\hline
		ULDor \cite{Tang2019Uldor}	& 69.22	&& --	& --	&& --	& -- \\
		3DCE \cite{Yan20183DCE}	& 75.55	&& --	& --	&& --	& -- \\
		LesaNet \cite{Yan2019Lesa}	& --	&& 93.98	& 43.44	&& --	& -- \\
		Auto RECIST \cite{Tang2018RECIST} & --	&& --	& --	&& --	& \bf 1.7088 \\
		MULAN	& \bf 85.22	&& \bf 95.12  & \bf 46.12	&& 1.4354 & 1.9619 \\
		\hline
	\end{tabular}
	\label{tbl:res_171tags} 
\end{table}

The detection performance of MULAN at various FPs per image is reported in Table \ref{tbl:acc_det}. The 171 tags were used for training, so the results are slightly different from those in Table 1 of the main paper.

\begin{table}[]
	\centering
	\setlength{\tabcolsep}{5pt}
	\renewcommand{\arraystretch}{1.2}
	\caption{Sensitivity (\%) at various FPs per image on the test set of DeepLesion (the 171 tags were used for training MULAN).}
	\begin{tabular}{lccccccc}
		\hline
		FPs per image	& 0.5	& 1		& 2		& 4		& 8		& 16 	& Avg.\ of [0.5,1,2,4] \\
		\hline
		3DCE \cite{Yan20183DCE}	& 62.48	& 73.37	& 80.70	& 85.65	& 89.09	& 91.06	& 75.55 \\
		ULDor \cite{Tang2019Uldor}	& 52.86	& 64.80	& 74.84	& 84.38	& 87.17	& 91.80	& 69.22 \\
		MULAN	& \bf 76.12	& \bf 83.69	& \bf 88.76	& \bf 92.30	& \bf 94.71	& \bf 95.64	& \bf 85.22 \\
		\hline
	\end{tabular}
	
	\label{tbl:acc_det}
\end{table}

Tables \ref{tbl:acc_per_tag1}--\ref{tbl:acc_per_tag3} show the details of the 171 tags and the tag-wise detection and tagging accuracies. The accuracies were computed on the mined tags \cite{Yan2019Lesa} of the validation set of DeepLesion. ``\# Train'' and ``\# Test'' are the numbers of positive cases in the training and validation sets, respectively.

\begin{table}[]
	\centering
	\scriptsize
	\setlength{\tabcolsep}{5pt}
	\renewcommand{\arraystretch}{1.2}
	\caption{Details of the 171 tags. The detection accuracy (average sensitivity in \%), tagging AUC and F1 (\%) are also shown.}
	\begin{tabular}{p{3.6cm}cccccc}
		\hline
		Tag	& Class	& \# Train	& \# Test	& Det.	& Tag AUC	& Tag F1 \\
		chest	&  bodypart	& 6813	& 784	& 76	& 94	& 72 \\
		abdomen	&  bodypart	& 5752	& 589	& 71	& 94	& 69 \\
		lymph node	&  bodypart	& 4752	& 516	& 74	& 95	& 67 \\
		lung	&  bodypart	& 3481	& 414	& 79	& 97	& 87 \\
		upper abdomen	&  bodypart	& 2957	& 347	& 71	& 95	& 66 \\
		retroperitoneum	&  bodypart	& 2340	& 241	& 70	& 97	& 71 \\
		liver	&  bodypart	& 1838	& 200	& 73	& 98	& 81 \\
		pelvis	&  bodypart	& 1634	& 176	& 68	& 98	& 79 \\
		mediastinum	&  bodypart	& 1604	& 170	& 77	& 96	& 61 \\
		right lung	&  bodypart	& 1579	& 195	& 75	& 97	& 71 \\
		left lung	&  bodypart	& 1306	& 157	& 81	& 99	& 82 \\
		mediastinum lymph node	&  bodypart	& 1152	& 130	& 77	& 97	& 66 \\
		kidney	&  bodypart	& 896	& 116	& 66	& 98	& 83 \\
		soft tissue	&  bodypart	& 851	& 82	& 63	& 82	& 28 \\
		hilum	&  bodypart	& 681	& 65	& 69	& 96	& 79 \\
		chest wall	&  bodypart	& 672	& 69	& 51	& 98	& 72 \\
		left lower lobe	&  bodypart	& 654	& 63	& 70	& 99	& 71 \\
		right lower lobe	&  bodypart	& 647	& 79	& 79	& 98	& 61 \\
		right upper lobe	&  bodypart	& 518	& 72	& 73	& 98	& 65 \\
		left upper lung	&  bodypart	& 512	& 79	& 85	& 99	& 72 \\
		abdomen lymph node	&  bodypart	& 504	& 44	& 61	& 92	& 43 \\
		axilla	&  bodypart	& 493	& 41	& 62	& 100	& 85 \\
		mesentery	&  bodypart	& 468	& 47	& 69	& 97	& 52 \\
		bone	&  bodypart	& 462	& 52	& 49	& 96	& 65 \\
		paraaortic	&  bodypart	& 462	& 40	& 74	& 98	& 46 \\
		pelvis lymph node	&  bodypart	& 439	& 67	& 72	& 98	& 72 \\
		retroperitoneum lymph node	&  bodypart	& 432	& 31	& 75	& 98	& 39 \\
		pancreas	&  bodypart	& 395	& 31	& 56	& 99	& 52 \\
		adrenal gland	&  bodypart	& 377	& 50	& 79	& 100	& 80 \\
		axilla lymph node	&  bodypart	& 346	& 23	& 64	& 99	& 68 \\
		blood vessel	&  bodypart	& 341	& 47	& 68	& 86	& 35 \\
		left kidney	&  bodypart	& 320	& 45	& 62	& 99	& 68 \\
		hilum lymph node	&  bodypart	& 318	& 28	& 73	& 99	& 68 \\
		right kidney	&  bodypart	& 278	& 41	& 73	& 99	& 67 \\
		pleura	&  bodypart	& 275	& 43	& 55	& 95	& 29 \\
		right mid lung	&  bodypart	& 257	& 32	& 68	& 96	& 57 \\
		groin	&  bodypart	& 226	& 32	& 66	& 98	& 68 \\
		pelvic wall	&  bodypart	& 222	& 24	& 65	& 97	& 45 \\
		left adrenal gland	&  bodypart	& 220	& 29	& 74	& 100	& 75 \\
		spleen	&  bodypart	& 208	& 23	& 70	& 95	& 67 \\
		spine	&  bodypart	& 176	& 23	& 63	& 95	& 57 \\
		neck	&  bodypart	& 171	& 16	& 72	& 99	& 45 \\
		mesentery lymph node	&  bodypart	& 170	& 20	& 59	& 96	& 38 \\
		iliac lymph node	&  bodypart	& 165	& 29	& 77	& 99	& 55 \\
		lung base	&  bodypart	& 164	& 20	& 71	& 95	& 23 \\
		muscle	&  bodypart	& 163	& 21	& 79	& 95	& 26 \\
		porta Hepatis	&  bodypart	& 162	& 16	& 50	& 94	& 41 \\
		right hilum lymph node	&  bodypart	& 159	& 13	& 77	& 100	& 63 \\
		groin lymph node	&  bodypart	& 144	& 28	& 65	& 98	& 67 \\
		fat	&  bodypart	& 140	& 21	& 70	& 87	& 11 \\
		subcarinal lymph node	&  bodypart	& 138	& 26	& 83	& 99	& 56 \\
		fissure	&  bodypart	& 128	& 13	& 75	& 96	& 18 \\
		body wall	&  bodypart	& 127	& 18	& 86	& 99	& 28 \\
		right adrenal gland	&  bodypart	& 124	& 20	& 84	& 100	& 67 \\
		subcutaneous	&  bodypart	& 121	& 34	& 76	& 98	& 43 \\
		lingula	&  bodypart	& 120	& 14	& 86	& 99	& 27 \\
		porta Hepatis lymph node	&  bodypart	& 119	& 11	& 50	& 93	& 36 \\

		\hline
	\end{tabular}
	
	\label{tbl:acc_per_tag1}
\end{table}

\begin{table}[]
	\centering
	\scriptsize
	\setlength{\tabcolsep}{5pt}
	\renewcommand{\arraystretch}{1.2}
	\caption{Table \ref{tbl:acc_per_tag1} continued.}
	\begin{tabular}{p{3.6cm}cccccc}
		\hline
		Tag	& Class	& \# Train	& \# Test	& Det.	& Tag AUC	& Tag F1 \\
		superior mediastinum	&  bodypart	& 116	& 8	& 72	& 96	& 11 \\
		peritoneum	&  bodypart	& 115	& 3	& 92	& 88	& 0 \\
		superior mediast. lymph node	&  bodypart	& 108	& 16	& 80	& 96	& 21 \\
		paraspinal	&  bodypart	& 107	& 11	& 77	& 91	& 23 \\
		external iliac lymph node	&  bodypart	& 103	& 15	& 73	& 99	& 38 \\
		supraclavicular lymph node	&  bodypart	& 102	& 9	& 86	& 96	& 40 \\
		breast	&  bodypart	& 101	& 24	& 70	& 95	& 46 \\
		thyroid gland	&  bodypart	& 100	& 12	& 69	& 100	& 55 \\
		aorticopulmonary window	&  bodypart	& 99	& 11	& 75	& 99	& 44 \\
		intestine	&  bodypart	& 97	& 9	& 61	& 94	& 9 \\
		anterior mediastinum	&  bodypart	& 92	& 8	& 91	& 99	& 37 \\
		pancreatic head	&  bodypart	& 89	& 10	& 35	& 99	& 30 \\
		common iliac lymph node	&  bodypart	& 84	& 8	& 53	& 98	& 17 \\
		abdominal wall	&  bodypart	& 83	& 7	& 79	& 100	& 33 \\
		left hilum lymph node	&  bodypart	& 83	& 13	& 71	& 100	& 63 \\
		extremity	&  bodypart	& 79	& 17	& 85	& 97	& 20 \\
		adnexa	&  bodypart	& 73	& 7	& 43	& 98	& 31 \\
		paracaval lymph node	&  bodypart	& 72	& 3	& 83	& 100	& 16 \\
		airway	&  bodypart	& 71	& 10	& 45	& 97	& 23 \\
		aorta	&  bodypart	& 71	& 5	& 45	& 92	& 0 \\
		cardiophrenic	&  bodypart	& 69	& 4	& 75	& 99	& 19 \\
		rib	&  bodypart	& 66	& 11	& 14	& 99	& 53 \\
		diaphragm	&  bodypart	& 64	& 11	& 59	& 89	& 0 \\
		pancreatic tail	&  bodypart	& 63	& 3	& 50	& 99	& 9 \\
		paraspinal muscle	&  bodypart	& 59	& 4	& 88	& 90	& 13 \\
		peripancreatic lymph node	&  bodypart	& 57	& 4	& 25	& 98	& 13 \\
		omentum	&  bodypart	& 55	& 7	& 46	& 96	& 5 \\
		thigh	&  bodypart	& 54	& 12	& 90	& 98	& 24 \\
		psoas muscle	&  bodypart	& 54	& 4	& 88	& 89	& 14 \\
		thoracic spine	&  bodypart	& 51	& 9	& 50	& 100	& 44 \\
		subpleural	&  bodypart	& 51	& 7	& 61	& 97	& 14 \\
		vertebral body	&  bodypart	& 50	& 8	& 41	& 100	& 46 \\
		retrocrural lymph node	&  bodypart	& 50	& 4	& 81	& 79	& 50 \\
		lumbar	&  bodypart	& 48	& 4	& 69	& 99	& 30 \\
		perihilar	&  bodypart	& 48	& 1	& 100	& 96	& 0 \\
		pretracheal lymph node	&  bodypart	& 47	& 10	& 83	& 97	& 16 \\
		bronchus	&  bodypart	& 47	& 9	& 56	& 98	& 15 \\
		small bowel	&  bodypart	& 46	& 8	& 66	& 96	& 7 \\
		anterior abdominal wall	&  bodypart	& 46	& 3	& 83	& 100	& 23 \\
		pancreatic body	&  bodypart	& 45	& 3	& 42	& 100	& 46 \\
		cervix	&  bodypart	& 43	& 0	&-	& 0	& 0 \\
		stomach	&  bodypart	& 40	& 4	& 50	& 94	& 8 \\
		urinary bladder	&  bodypart	& 40	& 6	& 63	& 99	& 17 \\
		lung apex	&  bodypart	& 36	& 5	& 90	& 98	& 9 \\
		sacrum	&  bodypart	& 33	& 3	& 75	& 100	& 46 \\
		gallbladder	&  bodypart	& 33	& 2	& 50	& 100	& 36 \\
		biliary system	&  bodypart	& 33	& 3	& 42	& 92	& 33 \\
		pelvic bone	&  bodypart	& 32	& 3	& 50	& 95	& 22 \\
		sternum	&  bodypart	& 31	& 3	& 0	& 100	& 26 \\
		skin	&  bodypart	& 31	& 8	& 41	& 89	& 21 \\
		pericardium	&  bodypart	& 29	& 4	& 63	& 98	& 7 \\
		right thyroid lobe	&  bodypart	& 28	& 3	& 75	& 100	& 29 \\
		femur	&  bodypart	& 25	& 1	& 50	& 97	& 0 \\
		cortex	&  bodypart	& 25	& 3	& 50	& 95	& 4 \\
		trachea	&  bodypart	& 23	& 2	& 25	& 99	& 33 \\
		ovary	&  bodypart	& 22	& 3	& 92	& 98	& 10 \\
		subcutaneous fat	&  bodypart	& 21	& 10	& 70	& 98	& 20 \\
		lesser sac	&  bodypart	& 15	& 2	& 50	& 99	& 0 \\
		\hline
\end{tabular}

\label{tbl:acc_per_tag2}
\end{table}

\begin{table}[]
	\centering
	\scriptsize
	\setlength{\tabcolsep}{5pt}
	\renewcommand{\arraystretch}{1.2}
	\caption{Table \ref{tbl:acc_per_tag2} continued.}
	\begin{tabular}{p{3.6cm}cccccc}
		\hline
		Tag	& Class	& \# Train	& \# Test	& Det.	& Tag AUC	& Tag F1 \\
		mass	&  type	& 4037	& 412	& 77	& 84	& 21 \\
		nodule	&  type	& 3336	& 403	& 76	& 89	& 53 \\
		enlargement	&  type	& 996	& 114	& 76	& 76	& 2 \\
		lung nodule	&  type	& 752	& 77	& 86	& 94	& 38 \\
		lymphadenopathy	&  type	& 739	& 79	& 75	& 87	& 17 \\
		cyst	&  type	& 584	& 83	& 74	& 89	& 35 \\
		opacity	&  type	& 323	& 37	& 64	& 94	& 24 \\
		lung mass	&  type	& 280	& 19	& 93	& 93	& 12 \\
		metastasis	&  type	& 267	& 21	& 80	& 82	& 13 \\
		fluid	&  type	& 258	& 29	& 59	& 91	& 22 \\
		cancer	&  type	& 250	& 14	& 82	& 72	& 4 \\
		ground-glass opacity	&  type	& 192	& 23	& 54	& 96	& 22 \\
		thickening	&  type	& 176	& 28	& 46	& 79	& 24 \\
		consolidation	&  type	& 165	& 15	& 62	& 99	& 31 \\
		liver mass	&  type	& 160	& 23	& 85	& 93	& 28 \\
		infiltrate	&  type	& 119	& 9	& 64	& 92	& 10 \\
		necrosis	&  type	& 111	& 11	& 75	& 91	& 26 \\
		hemangioma	&  type	& 84	& 8	& 66	& 95	& 4 \\
		solid pulmonary nodule	&  type	& 64	& 7	& 71	& 91	& 4 \\
		kidney cyst	&  type	& 53	& 3	& 83	& 98	& 6 \\
		scar	&  type	& 48	& 10	& 63	& 80	& 5 \\
		adenoma	&  type	& 32	& 4	& 94	& 99	& 14 \\
		implant	&  type	& 30	& 1	& 0	& 66	& 0 \\
		expansile	&  type	& 29	& 0	& 0	& 0	& 0 \\
		lobular mass	&  type	& 24	& 3	& 92	& 81	& 0 \\
		simple cyst	&  type	& 24	& 3	& 83	& 98	& 7 \\
		lipoma	&  type	& 21	& 3	& 0	& 99	& 8 \\
		hypoattenuation	&  attribute	& 1681	& 188	& 70	& 90	& 40 \\
		enhancing	&  attribute	& 902	& 120	& 66	& 81	& 19 \\
		large	&  attribute	& 558	& 46	& 77	& 83	& 11 \\
		prominent	&  attribute	& 396	& 37	& 64	& 91	& 18 \\
		calcified	&  attribute	& 375	& 45	& 71	& 75	& 3 \\
		indistinct	&  attribute	& 278	& 18	& 61	& 82	& 11 \\
		solid	&  attribute	& 267	& 28	& 56	& 87	& 21 \\
		hyperattenuation	&  attribute	& 239	& 30	& 72	& 83	& 21 \\
		heterogeneous	&  attribute	& 191	& 19	& 75	& 81	& 20 \\
		spiculated	&  attribute	& 170	& 26	& 70	& 79	& 23 \\
		sclerotic	&  attribute	& 168	& 20	& 54	& 99	& 58 \\
		soft tissue attenuation	&  attribute	& 147	& 12	& 60	& 69	& 5 \\
		tiny	&  attribute	& 126	& 21	& 67	& 93	& 11 \\
		lobular	&  attribute	& 102	& 18	& 85	& 68	& 3 \\
		conglomerate	&  attribute	& 98	& 20	& 78	& 84	& 16 \\
		lytic	&  attribute	& 90	& 11	& 20	& 99	& 31 \\
		cavitary	&  attribute	& 73	& 26	& 87	& 93	& 26 \\
		subcentimeter	&  attribute	& 72	& 11	& 86	& 89	& 7 \\
		circumscribed	&  attribute	& 70	& 2	& 50	& 63	& 0 \\
		diffuse	&  attribute	& 62	& 9	& 67	& 65	& 4 \\
		exophytic	&  attribute	& 59	& 9	& 72	& 85	& 12 \\
		oval	&  attribute	& 41	& 4	& 31	& 68	& 9 \\
		fat-containing	&  attribute	& 37	& 10	& 53	& 74	& 6 \\
		noncalcified	&  attribute	& 35	& 5	& 95	& 96	& 6 \\
		nonenhancing	&  attribute	& 34	& 8	& 72	& 88	& 4 \\
		lucent	&  attribute	& 33	& 3	& 75	& 8	& 0 \\
		thin	&  attribute	& 20	& 7	& 57	& 91	& 13 \\
		reticular	&  attribute	& 17	& 6	& 83	& 93	& 4 \\
		patchy	&  attribute	& 14	& 3	& 67	& 84	& 0 \\
		\hline
	\end{tabular}

\label{tbl:acc_per_tag3}
\end{table}
%
%
%
\bibliographystyle{splncs04}
\bibliography{paper354}

\end{document}